\documentclass{article}

\usepackage[english]{babel}

\usepackage[letterpaper,top=2cm,bottom=2cm,left=3cm,right=3cm,marginparwidth=1.75cm]{geometry}

\usepackage{amsmath, amssymb}
\usepackage{graphicx}
\usepackage{natbib}
\usepackage{multirow}
\usepackage{booktabs} 
\usepackage[colorlinks,linkcolor=blue,filecolor=blue,citecolor=magenta,urlcolor=blue]{hyperref}
\usepackage{algorithm}
\usepackage{algorithmic}
\usepackage{soul}
\usepackage{subcaption}
\usepackage{wrapfig}
\usepackage{bbm}
\usepackage{spverbatim}
\usepackage{inconsolata}
\usepackage[affil-it]{authblk}

\newtheorem{definition}{Definition}

\title{Retrieval Meets Reasoning: Dynamic In-Context Editing for Long-Text Understanding}

\date{\vspace{-5ex}}

\author[1,2]{Weizhi Fei}
\author[2]{Xueyan Niu\thanks{Correspondence to: \texttt{niuxueyan3@huawei.com}.}}
\author[3]{Guoqing Xie}
\author[3]{Yanhua Zhang}
\author[2]{Bo Bai}
\author[2]{Lei Deng}
\author[2]{Wei Han}

\affil[1]{%
    Department of Mathematical Sciences, Tsinghua University, Beijing, China}
\affil[2]{%
    Theory Lab, 2012 Labs, Huawei Technologies Co., Ltd.}
\affil[3]{%
    Architecture \& Design, ICT Products \& Solutions, Huawei Technologies Co., Ltd.}

\begin{document}
\maketitle

\begin{abstract}
Current Large Language Models (LLMs) face inherent limitations due to their pre-defined context lengths, which impede their capacity for multi-hop reasoning within extensive textual contexts. While existing techniques like Retrieval-Augmented Generation (RAG) have attempted to bridge this gap by sourcing external information, they fall short when direct answers are not readily available. We introduce a novel approach that re-imagines information retrieval through dynamic in-context editing, inspired by recent breakthroughs in knowledge editing. By treating lengthy contexts as malleable external knowledge, our method interactively gathers and integrates relevant information, thereby enabling LLMs to perform sophisticated reasoning steps.
Experimental results demonstrate that our method effectively empowers context-limited LLMs, such as Llama2, to engage in multi-hop reasoning with improved performance, which outperforms state-of-the-art context window extrapolation methods and even compares favorably to more advanced commercial long-context models. Our interactive method not only enhances reasoning capabilities but also mitigates the associated training and computational costs, making it a pragmatic solution for enhancing LLMs' reasoning within expansive contexts.

\end{abstract}

\section{Introduction}

\begin{figure*}[tb]
  \centering
  \includegraphics[width=.95\textwidth]{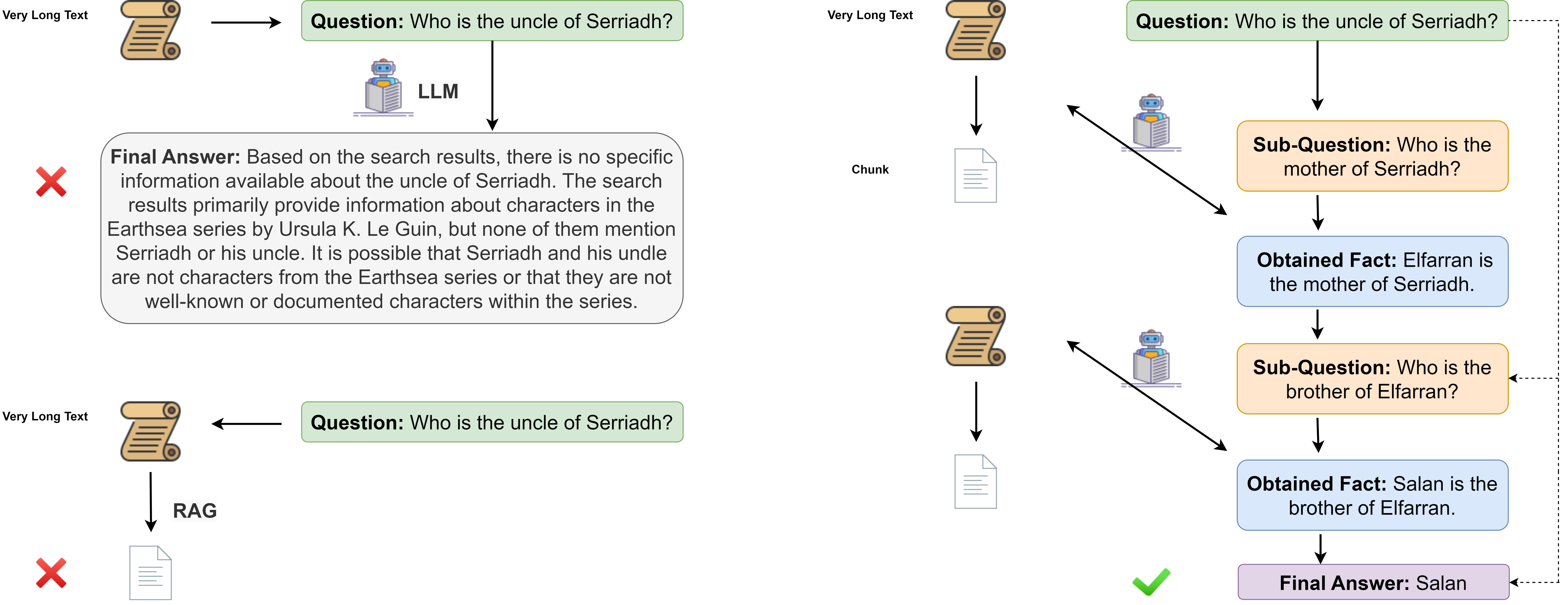}
  \caption{\small An instance of complex reasoning that involves synthesizing details from various parts of the text. As input context, the text sequence of \textit{The Earthsea Cycle} is provided to the model, and the task is to identify the uncle of the character Serriadh. \textbf{Top-left:} Using commercial LLM with longer context window, the final answer is coherent yet incorrect. \textbf{Bottom-left:} Using retrieval-augmented generation, the model is still unable to find the correct answer. \textbf{Right:} The proposed method interactively generates sub-questions and extracts relevant facts. These sub-questions and obtained facts are then used to plan subsequent steps. Given the input context's considerable length, the original text is segmented into manageable chunks. This allows the Large Language Model (LLM) to answer questions and retrieve information based on the most contextually relevant chunks.
  }
  \label{fig:method}
\end{figure*}

Recent Large Language Models (LLMs) have demonstrated remarkable performance in various natural language tasks. While these models have emerged with the ability to solve many real-world problems, their reasoning capabilities for interpreting complex prompts remain limited, particularly when the input context is lengthy \citep{hsieh2024ruler}. This limitation is due to the fixed context window inherent in current LLMs, which is often restricted to a pre-determined sequence length. For instance, both Meta's Llama2 \citep{touvron2023llama} and NVIDIA's Nemotron-4-340B-Base \citep{parmar2024nemotron} operate within a token limit of 4096, beyond which the models' reasoning abilities decline sharply. This issue is especially pronounced in tasks requiring long-range contextual reasoning, where capturing the full context is crucial for answering multi-hop questions. In Fig.~\ref{fig:method}, we illustrate an instance of this problem, showcasing the challenge of the reasoning task of deducing the uncle of the character Serriadh from the lengthy input context, \textit{The Earthsea Cycle} series by Ursula K. Le Guin, which contains approximately 407,495 words (and would translate to approximately 2M tokens). This question cannot be answered directly and requires synthesizing scattered pieces of information across the series.

Many techniques have been developed to extend the context window, including methods that require either modification of the model's layers \citep{munkhdalai2024leave} or fine-tuning \citep{chen2023longlora}, both of which incur substantial time and computational resources. Other approaches, such as semantic compression, aim to extract only relevant information from the long context \citep{fei2023extending}. While these methods are less costly and more efficient, they may falter in complex reasoning tasks that necessitate integration of information from the entire context. 
In fact, the top-left of Fig.~\ref{fig:method} shows the answer given by SoTA LLM, and the question remains unsolvable using SoTA LLM.
Retrieval-Augmented Generation (RAG) represents another line of research, designed to retrieve answers from long contexts. However, as depicted in the bottom-left of Fig.~\ref{fig:method}, using RAG directly does not yield the correct answer for questions like identifying Serriadh's uncle, as this information is not explicitly stated in the novels.

We propose a novel method that dynamically decomposes the question into sub-questions, forming a Directed Acyclic Graph (DAG), and leverages the model's reasoning ability to progress through the graph. The right-hand side of Fig.~\ref{fig:method} illustrates our approach, where the model interactively derives and addresses sub-questions, such as identifying the mother of Serriadh, by looking up relevant context chunks. This process continues interactively until the final answer is deduced. It has been demonstrated in \citep{huang2023large} that external feedback can significantly enhance the reasoning capabilities of models, underscoring the importance of exchanging information between the model and the provided context. Our method is inspired by in-context editing  \citep{zhong2023mquake,wang2024deepedit}, a technique that involves presenting the model with a set of instructions or examples along with the input text to guides the model's generation process in a targeted manner. This approach allows the model to focus on the external knowledge, thereby enhancing its hallucination. By incorporating elements of in-context editing, our method not only refines the model's interaction with the input but also fosters a more nuanced understanding of the underlying context. This results in a more robust and flexible reasoning framework that can effectively navigate the intricacies of long-form textual data, overcoming the limitations posed by traditional fixed-context window models.

Often, modifying the base model is inefficient and costly. Our approach provides a light-weight, plug-and-play solution for enhancing the reasoning capabilities of LLMs and can be easily incorporated
into other interpolation-based context
window extension methods and black box APIs. The proposed methods can be implemented on one single GPU while incurring no extra parameter updates or memory
consumption, making it a pragmatic solution for a wide range of applications.

\section{Problem Formulation}
\subsection{Reasoning over long context}

Let $C$ be the input context that consists of a sequence of words. In practice, those very long sequences is particular useful and attracted~\citep{Pawar2024TheWW}, such as scientific papers, novels, and legal contracts. The task of reasoning over long context often involves answering questions that requires reasoning and utilize relevant facts from multiple parts of $C$. We denote the correct answers of the question $Q$ as $\mathsf{Answer}(Q)$ and the reasoning result using language model $f$ given context $C$ as $f(Q|C).$ Specifically, given a multi-hop question $Q$, the answer $\mathsf{Answer}(Q)$ of $Q$ is obtained from the inter-dependent reasoning steps $(s_1, s_2, \ldots, s_t)$, where each reasoning step $s_i$ must adhere to the knowledge contained within the  context $C$. Usually, these reasoning steps and the dependencies can be represented by a Directed Acyclic Graph (DAG) known as the reasoning graph \citep{trivedi-etal-2022-musique}.
\begin{definition}
 The DAG $G_Q = (V, E)$ is the reasoning graph of the multi-hop question $Q$ if the nodes $s_i\in V$ correspond to the reasoning steps and the edges $(s_i, s_j) \in E$ indicate that the step $s_j$ relies critically on the output of the predecessor step $s_i$.
\end{definition}

In current LLMs, there is a maximum number of tokens, known as the context window, that each model is able to handle at one time. Many techniques have been developed to extend the context length of foundation LLMs (see Section~\ref{sec:related-work}). Using these methods, LLMs have already emerged the abilities to achieve high benchmark scores across a wide range of downstream tasks. However, when it comes to query answering systems~\citep{Yang2024CRAGC}, they still suffer from hallucination, a phenomenon where models generate plausible but incorrect answer that is unfaithful to the given context. Thus, models with extended lengths still result in low accuracy in complex reasoning tasks that involves long context, rendering $f(Q|C)\not\in \mathsf{Answer}(Q)$. Instead of extending the context window directly, in this paper, we view the provided context as a source of external knowledge and conduct inference along the reasoning DAG, such that at each step, adhering the available context is within the window $L$ of the LLM. In the following Section~\ref{sec: problem_ke}, we will present the target process is similar to the in context editing methods for multi-hop question.

Usually, the question $Q$ is very short (i.e., $n_Q \ll L$), but the length of the external context $C=(x_1,x_2,\ldots,x_{n_C})$ that can be used to assist the reasoning is very long, such that $n_C > L.$ For example, the question ``Who is the uncle of Serriadh?'' in Fig.~\ref{fig:method} consists of 9 tokens using GPT-2 tokenizer. However, the input context available to us, i.e., the text of \textit{The Earthsea Cycle} series, has about has 407,495 words (about 2M tokens), so $n_C\gg L$ for popular models such as Llama2 and Nemotron4.
In the sequel, we design algorithms for deriving $\mathsf{Answer}(Q)$ of multi-hop question $Q$ given long input $C> L.$

\subsection{Knowledge editing involving multi-hop reasoning}
\label{sec: problem_ke}

Let $\mathcal{E} = \{e_1, \cdots, e_{n_e}\}$ be a set of fact edits, where each edit $e_i$ is an updated fact $((s, r, o) \rightarrow (s, r, o^*)),$ where the tuple consisting of a subject ($s$), relation ($s$), and object ($o$) stands for the knowledge represented by relation fact~\citep{zhong2023mquake}.  Given a language model $f\in \mathcal{F}$, the task of knowledge editing is a function $K: \mathcal{F} \times \mathcal{E} \rightarrow \mathcal{F}$, such that the updated model $f_{\mathcal{E}}^* = K(f, \mathcal{E})\in \mathcal{F}$ incorporates the new knowledge in $\mathcal{E}$ when answering related questions.

To answer multi-hop questions relevant to edited knowledge~\citep{zhong2023mquake}, recent in-context editing methods ~\citep{zhong2023mquake,wang2024deepedit} generate the necessary steps and use retrieved edited knowledge to refine these steps within the context window. This success stems from LLMs' ability to understand instructions and follow new knowledge, as well as their reasoning ability to generate steps to solve complex problems. When answering multi-hop questions over long texts $C$, the challenge is that current LLMs are restricted by the context window and cannot fully understand the long text to get  the required knowledge. Therefore, reasoning over long texts is similar to knowledge editing involving multi-hop reasoning, if we can effectively retrieve the necessary knowledge from the unstructured long text.

As answering multi-hop questions accurately according to given context necessitates utilizing the knowledge within $C,$ 
we model the reasoning process with knowledge editing, namely, 
\begin{equation}\label{eq:ke}
f(Q|C) = f_C^*(Q) = K(f,\mathcal{E}_C)(Q),
\end{equation}
where $\mathcal{E}_C$ treat the context $C$ as the source containing the required new knowledge.

\section{Methodology}\label{sec:methodology}
Building upon the connection between multi-hop question answering over long contexts and knowledge editing involving multi-hop reasoning in Eq.~\eqref{eq:ke}, we apply multi-hop in-context editing \citep{zhong2023mquake,wang2024deepedit}, a technique that has been devised to execute knowledge editing, to our reasoning tasks using the long context as external knowledge.
Overall, our method includes two core modules: a planning module to generate the intermediate steps and a retrieval module to recall the relevant information from the given context to update the reasoning steps. 
We propose two methods for generating and updating intermediate steps, both of which incorporate planning and retrieval as integral components.

\subsection{Planning and retrieval}
\paragraph{Planning.}
The planning module leverage the reasoning abilities of LLMs to develop a step-by-step plan for complex tasks. This is crucial in addressing multi-hop question answering over long context for two main reasons: first, as demonstrated in \citep{wei2022chain}, solving complex tasks incrementally using Chain of Thought can significantly enhance the LLM's capability; second, the information required to answer a multi-hop question might be dispersed throughout the given context, in which case decomposing the question into sub-tasks is necessary for effective integration of pieces of information across the long text. As an example, consider the 2-hop question, ``Where was the wife of Francis I Rákóczi born?''. The intermediate information about the subject's wife is not provided in the question, neither does the base model have knowledge of such information. Direct retrieval from documents usually fails to recall the correct answer. By planning, the model is able to retrieve the necessary information for each intermediate step, thereby increasing the accuracy of the task.

Planning module attempt to decompose the complex question $Q$ into individual sub-task $q_1, \ldots, q_l$ that factorize into a DAG to facilitate the solution of the complex question $Q$. Specifically, the sub-task $q_i$ are obtained according to the original question $Q$ and the answers of the previous step $a_1, \ldots, a_{i-1},$ such that $q_i= D_{s+1}(Q, a_1, \ldots, a_{i-1}),$ where $a_j$ denote the answer of sub-question $q_j,$ and $D_{s+1}(\cdot)$ denotes the decomposition.
In practice, we leverage few-shot learning \citep{zhong2023mquake, wang2024deepedit} by providing successfully decomposed examples to guide the model in generating the next sub-question according to an appropriate reasoning graph and terminating the reasoning process. The provided examples ranging from chain reasoning steps to DAG reasoning steps can control the output style of the reasoning process. We present the example prompts of two algorithms in the Appendix~\ref{app: prompts}.

\paragraph{Retrieval.}
The retrieval module employs a sentence similarity model $\phi(\cdot, \cdot)$ to retrieve relevant information for a specified query. We consider both bi-encoder and cross-encoder models, where the bi-encoder generates sentence embeddings for semantic search within the embedding space, while the cross-encoder directly computes the similarity between sentence pairs. Although cross-encoder achieves better performance than bi-encoder, it is less efficient as is often the case. Typically, we first employ the bi-encoder to recall the top $k$ similar results then fine-grain the ranking using the cross-encoder.

As the external context $C$ is usually longer than the context window, in the retrieval module, we first divide $C$ into a set of chunks $\{c_1, c_2, \ldots, c_m\},$ where $c_i = (x_{i_1}, x_{i_2}, \ldots, x_{i_{n_i}}),\ i=1,2,\ldots, m$ such that $\cup_{i=1}^m \cup_{j=1}^{n_i} x_i = C,$ and the length of each sequence $n_i$ is less than $L.$ 
There are various methods for chunking and retrieval. For example, approaches in \cite{fei2023extending} can be applied. For long contexts, the text is divided into a list of chunks, each containing no more than $l\leq L$ words. The retrieval module then identifies the relevant chunks for query $q$ based on the similarity model $\phi(c_i, q)$. In practice, top-$k$ similar chunks $\{ c_q^1, \cdots, c_q^k \}$ from the long text usually contain the required knowledge of the sub-task. Consider the context order, we then concatenate these $k$ chunks, sorted by their original context indexes, into a single paragraph $C_q$. 
The required information for sub-question $q$ is likely to be in $C_q.$

\begin{figure}
    \begin{minipage}[c]{0.67\textwidth}
    \begin{subfigure}{0.4\textwidth}
        \includegraphics[width=\textwidth]{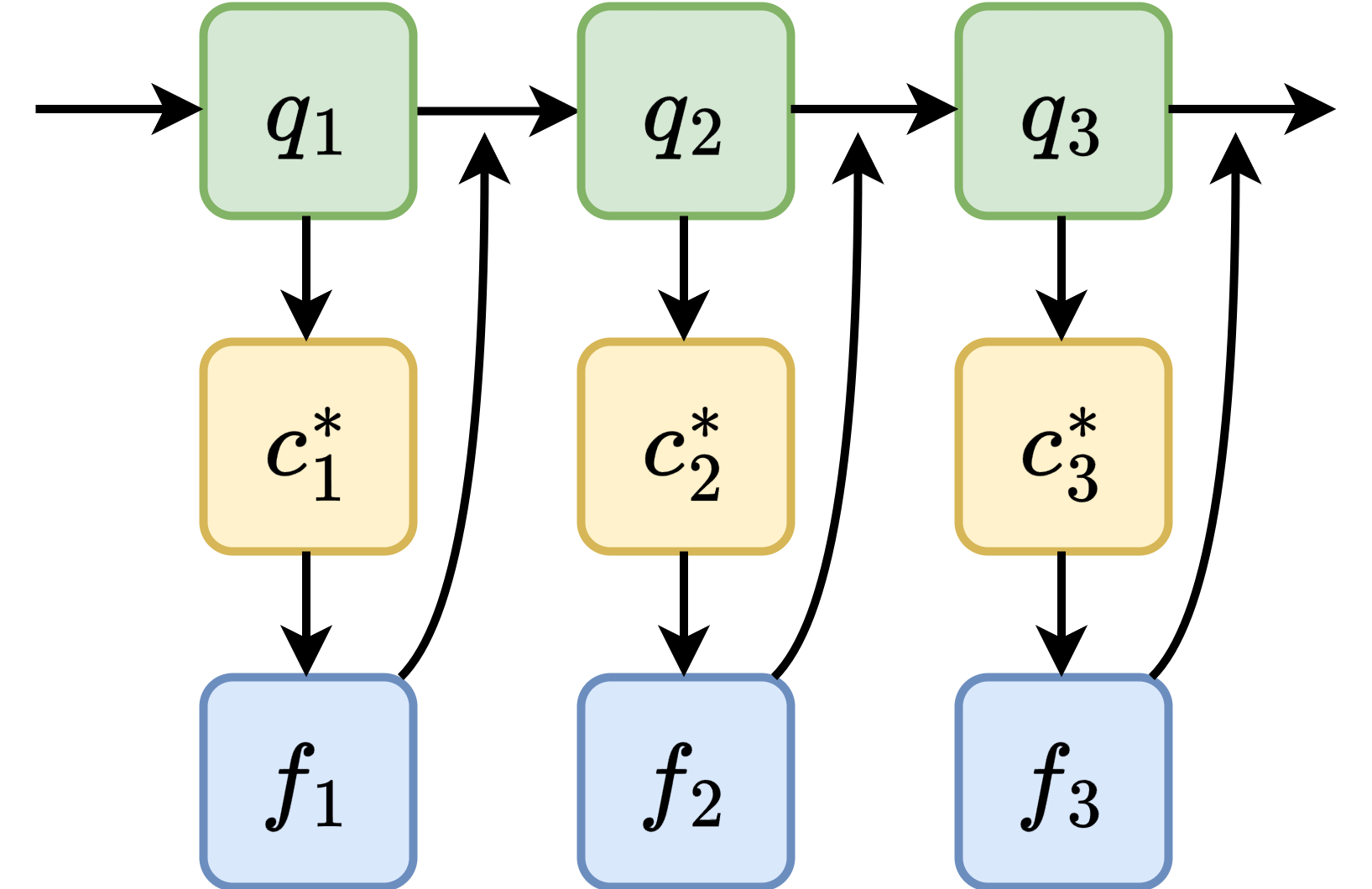}
        \caption{An illustration of Algorithm 1 on chains.}
        \label{fig:framework1}
    \end{subfigure}
    \hspace{1em}
    \begin{subfigure}{0.45\textwidth}
        \includegraphics[width=\textwidth]{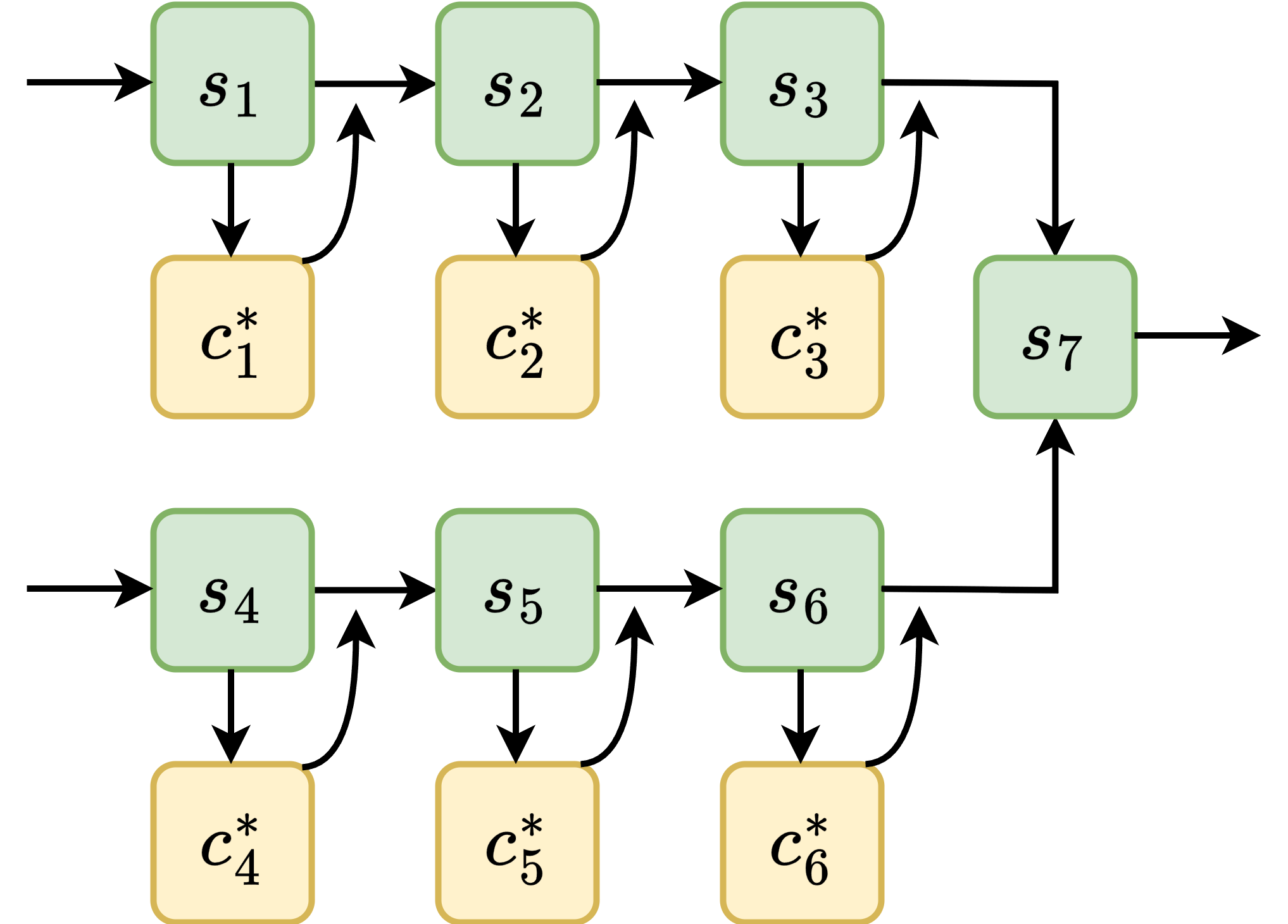}
        \caption{Alternative method using constrained decoding.}
        \label{fig:framework2}
    \end{subfigure}
    \end{minipage}\hfill
    \begin{minipage}[c]{0.3\textwidth}
    \caption{\small The proposed methods interactively utilize two standard modules: planning (depicted in green) and retrieval (depicted in yellow). As shown in illustration (a), a chain graph is employed; however, any arbitrary reasoning graph can be applied in general.}
    \label{fig: algithorm}
    \end{minipage}
\end{figure}

\begin{center}
\begin{minipage}{0.75\textwidth}
\begin{algorithm}[H]
\centering
\caption{Iterative QA With Fact Extraction}\label{alg:iterative}
\begin{algorithmic}
\REQUIRE Input a complex question $Q$, the list of chunks $\{c_i\}$ of the input context, a large language model LLM, a retrieval model, maximum iterations $M$, a decomposed task prompt $P_d$, and an extracted task prompt $P_e$.
\ENSURE Output the answer of $Q$
\STATE Initialize the reasoning thought $T \gets  Q$
\FOR{$i \gets 1$ to $M$}
\IF{$\text{LLM}(T)$ terminates the reasoning}
 \STATE Break
 \ELSE
 \STATE Generate Sub-question $q_i \gets Plan(P_d, T)$
 \ENDIF
 \STATE Get chunk $c^{*}_i \gets Retrieve(q_i, \{c_i\})$
 \STATE Obtain fact $f_i \gets \text{LLM}(P_e+q_i+c^*_i)$
 \STATE Update reasoning thoughts $T \gets T + q_i + f_i$
\ENDFOR

Obtain answer by reasoning thought $\text{LLM}(T)$
\end{algorithmic}
\end{algorithm}
\end{minipage}
\end{center}

\subsection{Iterative QA with fact extraction}
With the above modules, we proceed with our first algorithm using iterative QA~\citep{zhong2023mquake}, which generates a series of sub-question to solve the reasoning task in a step-by-step manner. 
As shown in Algorithm~\ref{alg:iterative},
the planing module first generates a series of sub-question $\{q_i\}$ using prompt $P_d$ which contains pre-defined successfully decomposed examples. The retrieval module then derives $c_i^* = C_q,$ which remains noisy with respect to the required knowledge. 

To reduce storage of the context window, after receiving $c_i^*$, we utilize an LLM to extract facts $f_i$ that can answer the sub-question $q_i$ according to the retrieved $C_{q_i}$. This process is commonly known as \textit{fact extraction}~\citep{agrawal-etal-2022-large} in NLP. In particular, we provide the LLM with prompt $P_e$ to extracts the facts that can answer the sub-question.

The proposed algorithm is illustrated in Fig.~\ref{fig:framework1} in the form of a chain graph.  Following the planned graph of sub-questions, the retrieval and fact extraction steps are repeated until the original complex task is solved. We aslo present thw whole process in Algorithm~\ref{alg:iterative}.

\subsection{Knowledge constrained decoding}\label{sec:method2}
We propose an alternative method that employs constrained decoding. Traditional constrained decoding modifies the probabilities of the vocabulary distribution during the decoding process. Inspired by the work of \citep{wang2024deepedit}, our approach circumvents accessing the distribution by directly decoding during the reasoning step. As depicted in Fig.~\ref{fig:framework2}, using example reasoning graphs, the LLMs formulate reasoning steps $s_i$ for a complex question during the planning stage and subsequently apply constrained decoding based on the retrieved $c_i^*$. This methodology ensures that the generated responses are both coherent and firmly grounded in the knowledge from the original context $C.$

The initially generated steps are intended as facilitators towards the solution, but they may occasionally include information that conflicts with the knowledge in the original context $C$. The retrieval module recalls relevant knowledge that can refine these statements. Examples of the prompts and instruction prompts are detailed in the Appendix~\ref{app: prompts}. The proposed algorithm is illustrated in Fig.~\ref{fig:framework2} in the form of a chain graph.

\section{Experiments}
We evaluate the reasoning capabilities of LLMs over long texts using our methods. We focus on the task of multi-document question answering from LongBench~\citep{bai2023longbench}, which involves multi-hop question answering over interconnected long texts. Additionally, we incorporate a multi-hop synthetic task from Ruler~\cite{hsieh2024ruler}, allowing us to control the sequence length of the text and the number of hops. Both tasks demand that the model assembles multiple pieces of information from the context and performs reasoning based on the evidence.

\subsection{Muti-hop question answering}
We investigate multi-document QA from an English benchmark LongBench~\cite{bai2023longbench} for long context understanding, which requires the model find and use information from several documents to solve the complex reasoning task. The datasets are tailored to the long-context setting, where the sequences that contains evidence for the multi-hop queries and the irrelevant sequences are randomly ordered to form the long input context. We utilize the F1 score from LongBench to quantify the similarity between the predicted answer and ground truth. The following datasets are used.

\begin{itemize}
    \item \textbf{HotpotQA} \citep{yang2018hotpotqa} is curated from Wikipedia that consists of numerous 2-hop questions crafted by native speakers. These questions are formulated based on two interconnected paragraphs and the average length of the context is 9,151 words. 
    \item \textbf{2WikiMultihopQA} \citep{ho-etal-2020-constructing} is also curated from Wikipedia, consisting of systematically constructed 5-hop questions using manual templates. Answering these questions requires reasoning across multiple paragraphs and cannot be solved by considering local content alone. On average, the length of the context in this dataset is 4,887 words.
    \item \textbf{MuSiQue} \citep{trivedi-etal-2022-musique} is curated from Wikipedia, specifically designed to include up to 4-hop questions while excluding shortcut and naturalness questions. Each question in this dataset is accompanied by 2-4 supplementary paragraphs that provide the reasoning path and relevant information. The average length of the context in this dataset is 11,214 words.
\end{itemize}

\subsection{Variable tracking}
Variable tracking is a multi-hop tracing synthetic task proposed in Ruler \citep{hsieh2024ruler} to emulate a minimal co-reference chain resolution task \citep{ng2010supervised}. Variable tracking requires the model to trace entities with multi-hop connections and find the target variable assignment chain within the long context. Specifically, a variable $X$ is initially assigned  with a number, followed by a linear sequence of variable assignments, forming a chain of variable name binding statements. Then these variable assignment will be randomly inserted into various positions within the padding text~\citep{mohtashami2023landmark}. The task complexity can be increased by adding more hops, more chains, and extending the length of the input.

\subsection{Baselines}

We consider LLMs, including the open-source Llama2-7B-chat-4k, as well as the commercial LLMs GPT-3.5-Turbo-16k and ChatGLM2-6B-32k. We also consider the following mainstream extension approaches for long context understanding as our baselines.

\paragraph{Fixed-size chunking.}
Chunking is a straightforward yet efficient approach to accommodate long contexts within a fixed-size context window. Since instructive information typically resides at the beginning and end of the sequence, the fixed-size chunking method \citep{bai2023longbench} truncates the input sequence from the middle when the input length exceeds the context window. The context window of the two commercial LLMs is large enough to cover most situations in the benchmark. However, the Llama2-7B model is still significantly restricted by the context window.

\paragraph{RAG.}
Retrieval Augmented Generation (RAG) models can be effectively applied to long context situations~\citep{xu2023retrieval}, where information relevant to a specific query can be retrieved from the given long context. RAG for long text understanding involves splitting the long context into smaller chunks of a default word size, and then employing a retriever to compute the embeddings of both the text chunks and the query. The top-$k$ chunks are selected based on their similarity to the query, and then fed as input to the model in order to generate the answer. We follow the implementation of RAG~\citep{bai2023longbench} and select the results where the Llama2-7B is used as the language model, text-embedding-ada-002 as the retrieval model, and the top 7 chunks of 200 words each are used.

\paragraph{Interpolation-based method.}

CLEX~\cite{chen2023extending} is a fine-tuning context window extrapolation method, which generalizes the position encoding (PE) scaling approaches to model the continuous dynamics using ordinary differential equations over the length scaling factor. CLEX extends the 4k context window of Llama2 to almost 8 times the training length. SelfExtend~\citep{jin2024llm} is a context window extension method that does not require any fine-tuning. It simply maps unseen relative positions into those seen during pre-training via the Floor operation. We select the SE-Llama2-7B 25k+ as a baseline, which is a modified version of the Llama2 with a 4k context window, extended to a 25k context window.

\section{Results}

\subsection{Mulit-hop question answering}
\begin{table}[tb]
\small
\centering
\begin{tabular}{ccccc}
\toprule
Model                  & HotpotQA & 2Wiki & MuSiQue & Avg   \\ \midrule
GPT-3.5-Turbo-16k      & 51.6     & 37.7     & 26.9  &38.7   \\ 
ChatGLM-6B-32k       & 45.1     & 34.0     & 21.9   & 33.6   \\ \midrule
 Llama2-7B-4k  \\
 \midrule
Fixed-size chunking     & 25.4     & 32.8     & 9.4  &25.6     \\ 
CLEX(k)      & 28.4     & 19.5     & 9.2  &19.0     \\ 
SE-Llama2-7B 25k+     & 35.5     & 30.5     & 15.5  & 27.2     \\ 
RAG      & 34.7     & 34.4     & 17.3   & 28.8    \\ 
Ours (Algorithm 1) & 44.8     & 48.0     & 34.1 & 42.3  \\ 
Ours (Algorithm 2) & \textbf{48.7}     & \textbf{50.3}     & \textbf{34.7} &  \textbf{44.5} \\
\bottomrule
\end{tabular}
\caption{\small Results of multi-hop question answering over long text for the three tasks,  HotpotQA, 2WikiMultiHopQA, and MuSiQue. The top rows are using commercial models with greater context windows. The bottom six rows are applying our methods and different baseline context extension methods.}\label{tab:longbench_multi}
\end{table}

We present the results of answering multi-hop questions over long context using different models and length extension methods in Table~\ref{tab:longbench_multi}. The results for GPT-3.5-Turbo-16k, ChatGLM-6B-32k, Llama2-7B (4k), and Llama2-7B (with retrieval) are sourced from LongBench \citep{bai2023longbench}. The results for CLEX(k) \citep{damonlpsg2023clex} and SE-Llama-7B 25k+ are taken from their respective papers.  For retrieval module of our algorithms, we consider the bge-base-en-v1.5 of BAAI.

Our experiments show that compared to extrapolation methods, retrieval methods can effectively enhance the ability of large models to use long texts to answer multi-hop questions, achieving performance competitive with commercial long-text models. Our approach, which iteratively retrieves and answers the texts needed for multi-hop questions, significantly outperforms direct retrieval-augmented methods and surpasses commercial long-text models on two tasks. Among our two approaches, the knowledge-constrained decoding method consistently outperforms the iterative questioning method across all three tasks. This is because iterative questioning, compared to direct generation steps, requires an additional step of converting the context into questions, which is sub-optimal. Our approach leverages the models' reasoning abilities to perform step-by-step inference, yielding better results with stronger models. More details can be found in Appendix~\ref{app: other results}.

\subsection{Multi-hop variable tracking}

The variable tracking task can be synthesized by considering the hop number, chain number, and sequence length. We generate 50 samples for each configuration. We compare our approach to the direct application of Llama2-7B on sequences ranging from 2k to 32k in length for these tasks. We conducted two experiments: the first investigates the impact of the number of hops by fixing the number of chains at 1 and varying the number of hops between 2, 4, and 8; the second explores the impact of the number of chains by fixing the number of hops at 2 and varying the number of chains between 1, 2, and 3. We evaluate the variable tracking task using the method outlined in Section~\ref{sec:method2}.

\begin{figure}[t]
    \centering
    \begin{subfigure}{0.45\textwidth}
        \includegraphics[width=\textwidth]{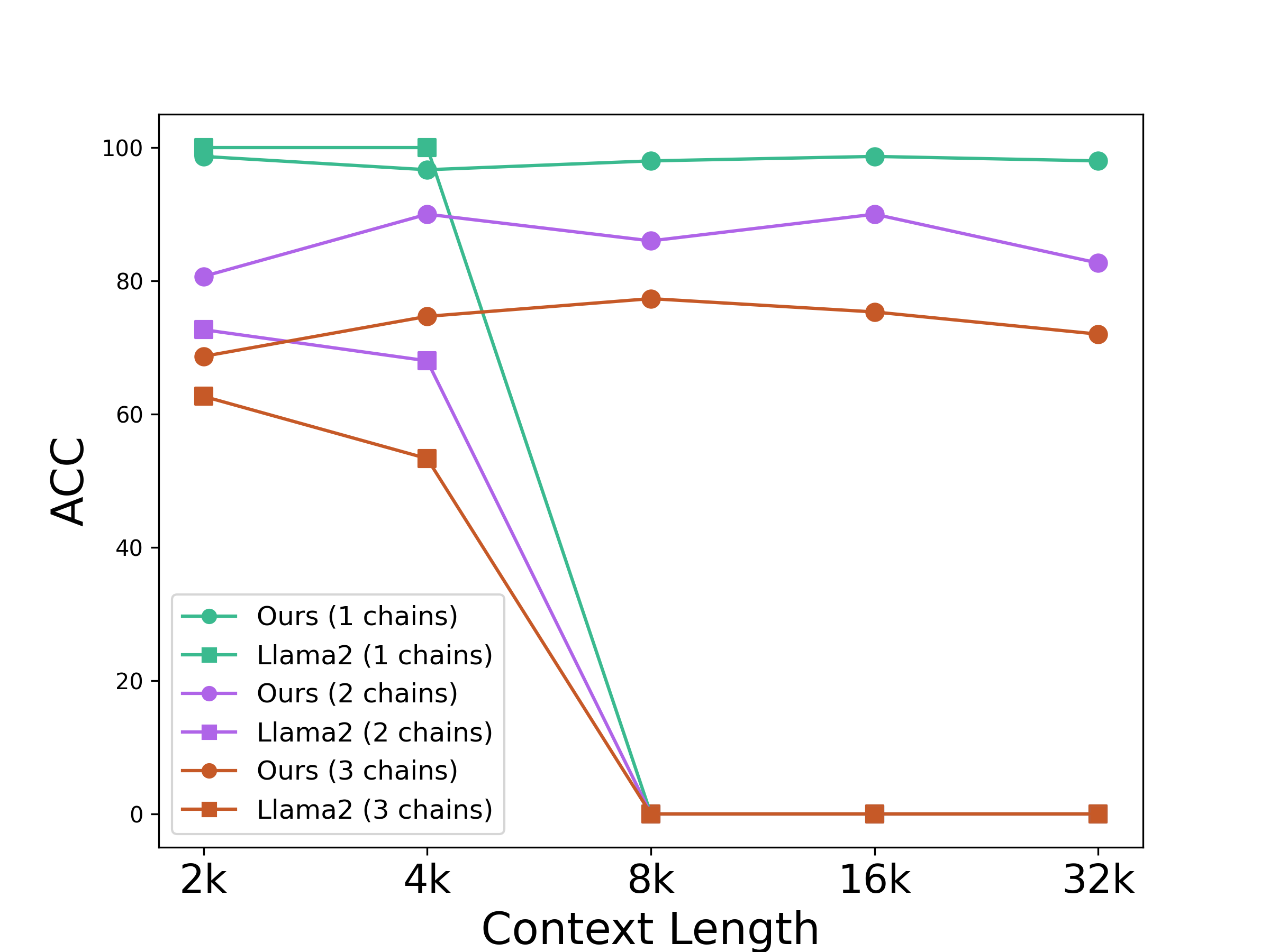}
        \caption{2-hop variable tracking}
        \label{fig: vt_chain}
    \end{subfigure}
    \begin{subfigure}{0.45\textwidth}
        \includegraphics[width=\textwidth]{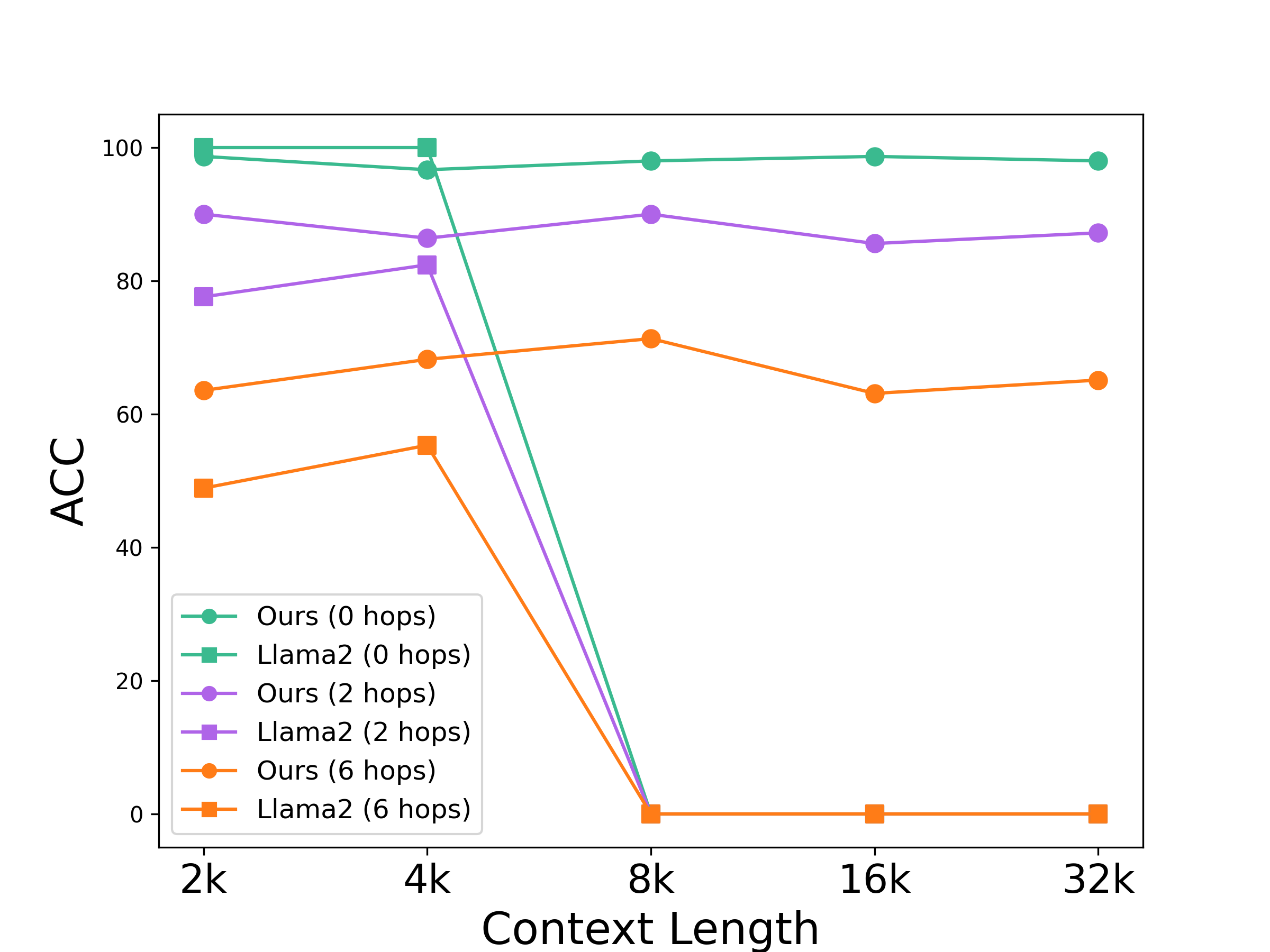}
        \caption{Variable tracking on a single chain}
        \label{fig: vt_hop}
    \end{subfigure}
    \caption{\small Variable tracking accuracies using Llama2 and our method. Our method, which utilizes knowledge-constrained decoding, is represented by squares, while Llama2 is represented by circles. The settings using different number of chains/hops are indicated by the color. \textbf{(a)} Varying the number of chains with the number of hops fixed at 2.  \textbf{(b)} Varying the number of hops with the number of chains fixed at 1. In all these settings, our methods maintain high accuracies.}
    \label{fig: vt}
\end{figure}

As depicted in Fig.~\ref{fig: vt}, across all configurations of Llama2, a rapid decline in accuracy to zero is observed once the input sequence exceeds the window size of 4096. However, utilizing our method, the accuracy remains largely unaffected even at the length 32k, underscoring the robustness of our approach to varying text lengths. The increase in the number of hops and chains introduces greater interference in the variable tracking process. Our approach constantly outperforms Llama2 in the majority of settings, even for sequences shorter than 4k, indicating its effectiveness in tracking multi-hop variable assignment chains.

\subsection{Ablation study}
\label{sec: resluts.ablation}

The planning and retrieval modules play pivotal roles in long-context reasoning. As larger models typically exhibit superior generalization capabilities, we verify the model parameter size for both the planning and retrieval modules and investigate the impact of enhancing their reasoning and retrieval capabilities. We adopt Llama2-7B and Llama2-13B as our base models. We utilize a series of English embedding models from BAAI for the retrieval module and apply the method outlined in Section~\ref{sec:method2} for our ablation study.

By altering the models within the planning and retrieval modules, we assess their average performance on multi-hop question answering tasks using the LongBench dataset, as depicted in Table~\ref{tab:ablation_longbench}. The findings indicate that performance generally improves as the size of the retrieval models increases. Specifically, an upgrade from Llama2-7B to Llama2-13B results in a significant performance boost, surpassing that of commercial long models. This suggests the potential of our approach when employing larger models endowed with more robust reasoning capabilities. We present the detailed findings of the ablation study in Appendix~\ref{app: other results}.

\begin{table}[t]
\centering
\small
\begin{tabular}{cccc}
\toprule
     Avg      & bge-small & bge-base & bge-large \\ \midrule

Llama2-7B  & 43.7  & 44.6 & 46.0  \\ \midrule
Llama2-13B &  48.7     &  50.6    & 50.3  \\ \bottomrule
\end{tabular}
\caption{\small Ablation study on the planing module and retrieval module. Bge-small, bge-base, bge-large represents BAAI bge-small-en-v1.5 (33M parameters), BAAI bge-base-en-v1.5 (109M parameters), and BAAI bge-large-en-v1.5 (335M parameters) respectively. We report the average scores of the three multi-hop question answering datasets.}\label{tab:ablation_longbench}
\end{table}

\section{Related Work}\label{sec:related-work}

\subsection{Knowledge editing}
\label{sec: related_ke}
Knowledge editing involves introducing new knowledge into static models, and multiple methods have been explored to modify relevant parameters to handle this task. Some approaches focus on locating and modifying model weights associated with specific knowledge~\citep{meng2022locating}, as well as rapid adaptation facilitated by a compact auxiliary editing network~\citep{de2021editing}. Notably, the emergent in-context learning method can be used to edit factual knowledge without updating parameters~\citep{zheng2023can}. \citet{zhong2023mquake} demonstrate that while previous knowledge-editing approaches that modify model parameters can accurately recall edited facts, they fail catastrophically on constructed multi-hop questions. Therefore, a multi-hop knowledge editing benchmark, MQUAKE~\citep{zhong2023mquake}, was proposed. Many in-context editing methods employ step-by-step reasoning to address multi-hop situations, including memory-based iterative query answering and constrained decoding~\citep{wang2024deepedit}.

\subsection{Reasoning}
\citet{wei2022chain} discover that incorporating intermediate steps in generation significantly improves the reasoning ability of LLMs and proposed the Chain-of-Thought (CoT) technique. Building upon CoT, \citet{khot2022decomposed} introduce a prompting approach to address complex problems by decomposing them into simpler sub-tasks. Subsequently, \citet{trivedi-etal-2023-interleaving} combine the decomposed prompting strategy with a retrieval approach, proposing iterative retrieval for solving open-domain complex question answering. Additionally, \citet{zhong2023mquake} integrate conflict remediation into iterative retrieval to tackle multi-hop question answering in knowledge editing.
Numerous studies have addressed the scope of reasoning plans. \citet{yao2023tree} propose the Tree of Thought (ToT) technique, which considers multiple reasoning paths and employs self-evaluation to determine the subsequent course of action. Additionally, \citet{besta2023graph} enhance the capacity to model information generated by LLMs by representing it as a flexible graph structure, enabling the combination of diverse LLM thoughts into synergistic outcomes and capturing the essence of interconnected thought networks. There is also research that considers reasoning by investigating the answering of complex queries over knowledge graphs~\cite{wang_logical_2023,wang_wasserstein-fisher-rao_2023}. These studies involve answering first-order logic queries~\cite{wang_benchmarking_2021} and extend beyond tree structures~\cite{yin2024rethinking} and single-variable queries~\cite{yin2024rethinking}.

\subsection{Retrieval}
Retrieval-Augmented Generation (RAG) has shown the potential to equivalently extend the context window of large language models for downstream tasks. \citet{xu2023retrieval} demonstrate that simple retrieval augmentation at generation can achieve comparable performance to fine-tuned LLMs on long-context tasks. Furthermore, \citet{zhao2024longagent} introduce LongAgent, inspired by multi-agent collaboration, which successfully extend the context window of LLaMA to 128K for multi-hop question answering tasks. \citet{sarthi2024raptor} propose RAPTOR, which organizes the retrieval into a tree-structure based on the input document. However, these methods directly retrieve based on the input query, often failing to obtain relevant information regarding the intermediate variables required for reasoning.

\subsection{Lossless extrapolation} 
A significant body of research focuses on adapting existing LLMs trained on short texts to accommodate longer texts during inference \citep{anil2022exploring}. The key challenge lies in the position encoding of the input, which has only been trained on short texts and is therefore inadequate for handling long texts. Current studies usually rely on relative positional encoding, such as Rotary Position Embedding (RoPE)~\citep{su2024roformer}, which are widely adopted due to their strong extrapolation capabilities. \citet{chen2023extending} develop the Position Interpolation (PI) method to linearly scale the positional encoding of long text into trained encoding. 
\citet{peng2023yarn} introduce YaRN, an efficient extrapolation mechanism using the neural tangent kernel, which dynamically scales the logits. \citet{damonlpsg2023clex} further propose continuous dynamics and utilized ordinary differential equations to fit the length scaling factor.

\section{Conclusion}
Reasoning has become a critical demand in the era of large language models (LLMs). Due to the constrained of LLMs on their  context lengths, various techniques have been proposed to extend these limits. Our study illuminates that the input context essentially serves as external knowledge that LLMs can access interactively to conduct inference. From this perspective, reasoning over long contexts is essentially equivalent to knowledge editing, which has been extensively studied. We propose two methods inspired by knowledge editing to enable LLMs with limited context windows to plan reasoning steps and retrieve relevant context. Experimental results demonstrate that our proposed methods outperform current approaches on the long-context multi-hop question answering task in LongBench as well as on the variable tracking task, which can be further improved with models that have better reasoning and retrieval abilities. We provide the implementation details and running time in Appendix~\ref{app: imple results}.
We present a discussion concerning the   limitations and potential risks of our study in Appendix~\ref{app: limit} and Appendix~\ref{app: risk}, respectively.

\bibliographystyle{acl_natbib}
\bibliography{ref}


\appendix

\section{Discussions}
\subsection{Limitation}\label{app: limit}
We employ in-context edit methods for multi-hop question answering to handle reasoning over long context tasks. However, due to the limitations of the available datasets, we only conduct our experiments on a subset of LongBench and Ruler. Additionally, the reasoning steps prompted by the provided example prompts should follow a specific instructed style, where the algorithm needs to execute based on that style, which may not be stable or generalizable. The instructions for the prompts may not be applicable to all large language models (LLMs). It would be an interesting direction to explore how to design a method that can generally generate robust reasoning steps.

\subsection{Potential risks}
\label{app: risk}
As a RAG-based context window extension approach for large language models, our proposed reasoning method enables the model to process longer texts. The experiments and evaluations we conducted utilized publicly available academic datasets, thus avoiding direct ethical concerns related to the use of private or sensitive data.
However, it is worth noting that our method could potentially be used to extract inferred private information from long texts in commercial settings. This is a common ethical concern associated with the use of long text processing models. For example, our approach could potentially be applied to extract personal details or sensitive information from customer service logs, medical records, or other long-form business documents, which could raise privacy issues if not handled carefully.
To mitigate these ethical risks, it would be important to clearly define the intended use cases for our method and implement appropriate safeguards, such as data anonymization, access controls, and transparency measures. 

\section{Supplementary Material}
\label{app: other results}
We present more experimental results of our Algorithm 2 in Table~\ref{tab: other_longbench_multi_2}. Since the baseline methods have not published their results on the Llama2-13B model, we only list the results of our method. The experimental results show that our method can achieve better performance under stronger retrieval models and language models. In the results on Llama2-13B, we can even comprehensively outperform the performance of GPT-3.5-Turbo-16k on three datasets.

\begin{table*}[tb]
\centering
\caption{Results of multi-hop question answering over long text for the three tasks,  HotpotQA, 2WikiMultiHopQA, and MuSiQue. The top rows are using commercial models with greater context windows. The middle six rows are applying our methods and different baseline context extension methods on Llama2-7B. The bottom rows are applying on on Llama2-13B. 7B and 13B represents the Llama2-7B-4k and Llama2-1B-4k respectively.  Small, base, large represents BAAI bge-small-en-v1.5 (33M parameters), BAAI bge-base-en-v1.5 (109M parameters), and BAAI bge-large-en-v1.5 (335M parameters) respectively.}
\label{tab: other_longbench_multi_2}
\begin{tabular}{ccccc}
\toprule
Model                  & HotpotQA & 2Wiki & MuSiQue & Avg   \\ \midrule
GPT-3.5-Turbo-16k      & 51.6     & 37.7     & 26.9  &38.7   \\ 
ChatGLM-6B-32k       & 45.1     & 34.0     & 21.9   & 33.6   \\ \midrule
 Llama2-7B-4k  \\
 \midrule
Fixed-size chunking     & 25.4     & 32.8     & 9.4  &25.6     \\ 
CLEX(k)      & 28.4     & 19.5     & 9.2  &19.0     \\ 
SE-Llama2-7B 25k+     & 35.5     & 30.5     & 15.5  & 27.2     \\ 
RAG      & 34.7     & 34.4     & 17.3   & 28.8    \\ 
Ours (Algorithm 2, 7B, small) & 46.7     & 50.5    & 34.0 &  43.7 \\
Ours (Algorithm 2, 7B, base) & 48.7     & 50.3    & 34.7 &  44.5 \\
Ours (Algorithm 2, 7B, large) & 49.4     & 52.4    & 36.3 &  46.0 \\ \midrule
 Llama2-13B-4k  \\
 \midrule
Ours (Algorithm 2, 13B, small) & 51.0     & 56.3     & 39.1 &  48.8 \\
Ours (Algorithm 2, 13B, base) & 53.1     & 60.7     & 38.1 &  50.6 \\
Ours (Algorithm 2, 13B, bge) & 50.2     & 59.4     & 41.4 &  50.3 \\

\bottomrule
\end{tabular}
\end{table*}

\section{Implementation Details}
\label{app: imple results}
Our proposed methods employ dynamic retrieval to enable the reasoning ability over long context of large language model, involving language model and retrieval model. For language model, we consider the popular open-source Llama2-7B and Llama2 13 B, where are available at: \url{https://huggingface.co/meta-llama/Llama-2-7b-chat-hf}. For retrieval model, we consider the popular open-source bge-small-en-v1.5, bge-base-en-v1.5, and bge-large-en-v1.5, where are available at: \url{https://huggingface.co/BAAI/bge-small-en-v1.5}. The experiment of Llama2-7B can be conducted on a single V100 GPU with 32GB of memory, but the Llama2-13B should be run on the  GPU with larger memory. We present the computational budget based on Llama2-13B in Table~\ref{tab: time}, where this results are independently run on a single A40 GPU with 48G memory. The word number of  chunk size is 80 and we recall top 3 chunks for multi-hop question answering. 
\begin{table}[]
\centering
\begin{tabular}{cccc}
\toprule
      Time (min)            & Hot  & 2Wiki & MuSi\\ \midrule
(Llama2-13B, bge-small) & 34 & 37    & 32 \\ \midrule
(Llama2-13B, bge-base)           &  35    & 38    & 32 \\ \midrule
(Llama2-13B, bge-large) &  37    & 38    & 35 \\ \bottomrule
\end{tabular}
    \caption{The average running time (minute) on multi-hop question answering dataset of LongBench. Bge-small, bge-base, bge-large represents BAAI bge-small-en-v1.5 (33M parameters), BAAI bge-base-en-v1.5 (109M parameters), and BAAI bge-large-en-v1.5 (335M parameters) respectively.}
    \label{tab: time}
\end{table}

\section{Example Prompts}
\label{app: prompts}

\begin{figure}
    \centering
    \begin{subfigure}{0.49\textwidth}
        \includegraphics[width=\textwidth]{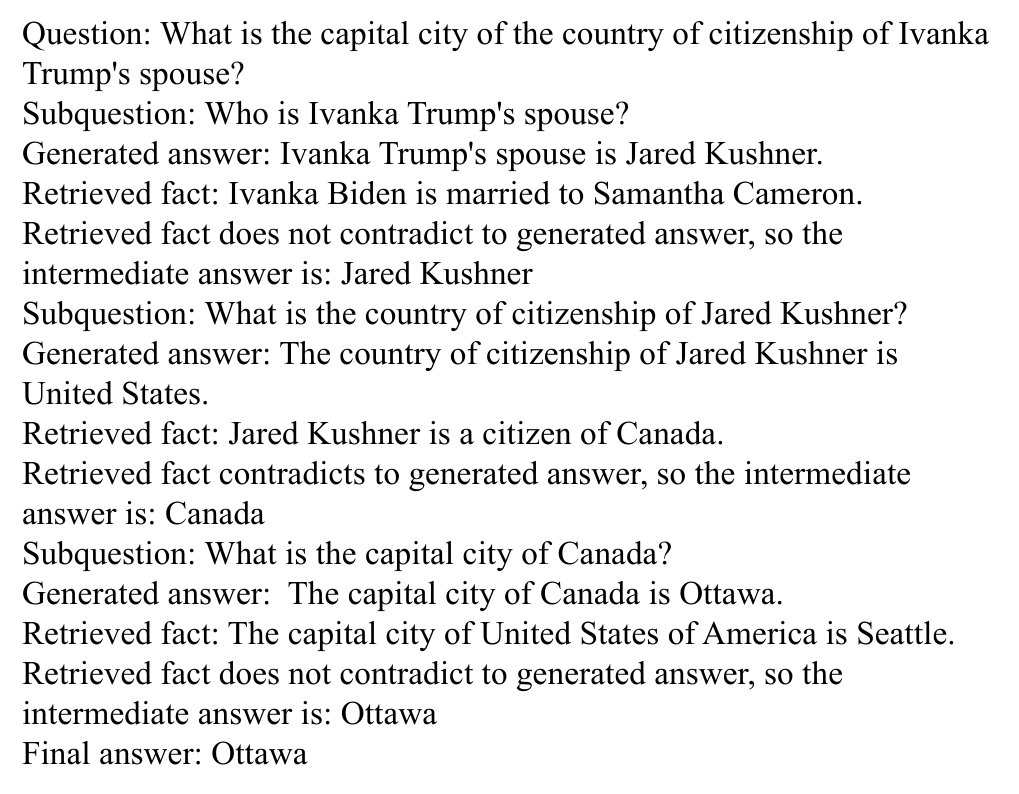}
        \caption{template example of algorithm 1}
        \label{fig: template prompt 1}
    \end{subfigure}
    \begin{subfigure}{0.49\textwidth}
        \includegraphics[width=\textwidth]{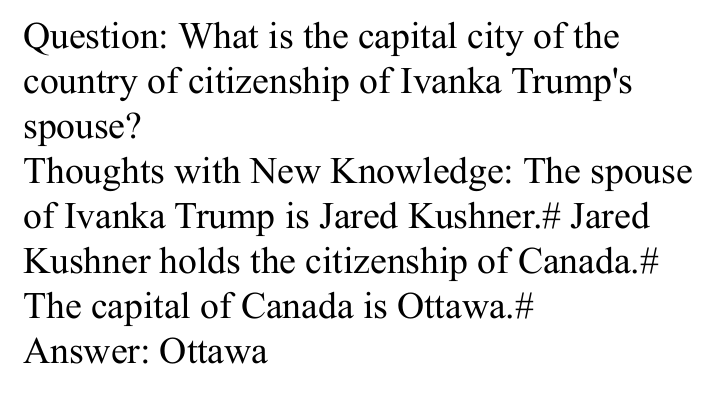}
        \caption{template  example of algorithm 2}
        \label{fig: template prompt 2}
    \end{subfigure}
    \caption{\small The template prompts of the two algorithms on the question ``What is the capital city of the country of citizenship of Ivanka Trump's spouse?''. \textbf{(a)} The template prompt of Algorithm 1, Iterative QA With Fact Extraction  \textbf{(b)} The template prompt of Algorithm 2, Knowledge constrained Decoding}
    \label{fig: template prompts}
\end{figure}

For the two algorithms, we selected 5 questions as examples, including 3 chain-style questions and 2 directed acyclic graph (DAG)-style questions, with the specific questions as follows. 

\begin{verbatim}
Question: What is the birth day of Rain Wilson' father?

Question: Who is the paternal grandfather of John Smith?

Question: What is the capital city of the country of citizenship of Ivanka Trump's spouse?

Question: Do both Django Unchained and Rango films have the directors from the same country?

Question: Which film has the director who died first, Love in the AfterNoon or Gigi?

\end{verbatim} .
The templates for each algorithm vary, and we selected one example with two different formatting styles, as shown in Fig.~\ref{fig: template prompts}. We executed the algorithms based on the given formats and decided when to terminate the process.

The prompt used for fact extraction is following:

\begin{verbatim}
Reference: {Context_Q}. Based on the reference, present the fact to answer the following question. 

Be concise, better one sentence. 

Question: {Question}. Fact:
\end{verbatim} 

\end{document}